\documentstyle[mkstyle,twoside]{article}
\title{Learning Belief Network Structure From Data under Causal 
Insufficiency
} 
\shorttitle{Belief Network under Causal Insufficiency}
\newcommand{\Bem}[1]{}

\date{}
\begin{document}

\machetitel

%\begin{abstract}
% \end{abstract}

\section{Introduction}

Various expert systems, dealing with uncertain data and knowledge, 
possess 
knowledge representation in terms of a 
belief network (e.g. knowledge base of the MUNIM system 
\cite{Andreassen:87} , ALARM network 
\cite{Cooper:92} etc.).
A 
number of  efficient algorithms for propagation of uncertainty within 
belief networks and their derivatives 
 have been developed,  compare e.g. \cite{Pearl:88}, \cite{Shachter:90b}, 
\cite{Shenoy:90}. 

Belief networks, causal networks, or influence diagrams, or (in Polish) 
cause-effect networks are terms frequently used interchangeably. They 
are quite 
popular for expressing causal relations under multiple variable setting 
both for deterministic and non-deterministic (e.g. stochastic) relationships 
in various 
domains: statistics, philosophy, artificial intelligence \cite{Geiger:90}, 
\cite{Spirtes:90b}. Though a belief network (a representation of the joint 
probability 
distribution, see \cite{Geiger:90}) and a causal network (a representation 
of 
causal relationships \cite{Spirtes:90b}) are intended to mean different 
things, 
they are closely related. Both assume an underlying dag (directed acyclic 
graph) structure of relations among variables and if Markov condition and 
faithfulness condition \cite{Spirtes:93} are met, then a causal network is in 
fact a belief network. The difference comes to appearance when we recover 
belief network and causal network structure from data. A dag of a belief 
network is satisfactory if the generated probability distribution fits the 
data, may be some sort of minimality is required. A causal network structure 
may be impossible to recover completely from data as not all directions of 
causal links may be uniquely determined \cite{Spirtes:93}. Fortunately, if  
we deal with causally sufficient sets of variables (that is whenever 
  significant influence variables are not omitted from observation), then 
there 
exists the possibility to identify the family of belief networks a causal 
network belongs to \cite{Verma:Pearl:90}. %\\

Regrettably, to our knowledge, a similar result is not directly known for 
causally insufficient sets of variables (that is when significant influence 
variables are hidden) - "Statistical indistinguishability is less well 
understood when graphs can contain variables representing unmeasured common 
causes" (\cite{Spirtes:93}, p. 88). Latent (hidden) variable 
identification has been investigated intensely both for belief networks (e.g. 
\cite{Pearl:86h}, \cite{Golmard:Mallet:89}, \cite{Liu:Wilkins:Ying:Bian:90}, 
\cite{Cooper:92}) and causal networks 
(\cite{{Pearl:Verma:91}}, \cite{Spirtes:90b}, \cite{Spirtes:93},
\cite{Glymour:87}, \cite{Glymour:Spirtes:88}), as well as 
in traditional statistics (see \cite{Spirtes:90c} for a comparative study of
LISREL and EQS techniques).
The algorithm of \cite{Cooper:92} recovers the most probable location of a 
hidden variable. Whereas the CI algorithm of \cite{Spirtes:93} recovers exact 
locations of common causes, but clearly not all of them. In fact, the CI 
algorithm does not provide a dag, but rather a graph with edges fully 
(unidirected or bidirected) or 
partially oriented, or totally non-oriented with additional constraints for 
 edge directions at other edges.  Partially or non-oriented edges  may prove 
to 
be either directed or bidirected edges. \\
The big open question is whether or not the bidirectional edges (that is 
indications of a common cause) are the only ones necessary to develop a belief 

network out of the product of CI, or must there be some other hidden 
variables added (e.g. by  guessing).  This  paper  is  devoted  to 
settling this 
question. %\\

\section{Causal Inference Algorithm}
%--------------------------\\

Below we remind the Causal Inference (CI) algorithm of Spirtes, Glymour and 
Scheines \cite{Spirtes:93} together with some basic notation used therein. 

Essentially, the CI algorithm recovers partially the structure of 
an including 
path graph. Given a directed acyclic graph G with the set of hidden nodes  
$V_h$ 
and visible nodes $V_s$ representing a causal network CN, an including path 
between nodes A and  B belonging to $V_s$ is a path in the graph G such that  
the only visible nodes (except for A and B) on the path are those where edges 
of the path meet head-to-head and there exists a directed  path  in G from 
such a node 
to either A or B. An including path graph for G is such a graph over $V_s$ in 
which if nodes A and B are connected by an including path in G ingoing into A 
and B, then A and B are connected by a bidirectional edge $A<->B$. Otherwise 
if they are connected by an including path in G outgoing from A and ingoing 
into B then A and B are connected by an unidirectional edge $A->B$.  % \\

A partially oriented including path graph contains the following types of 
edges unidirectional: $A->B$, bidirectional $A<->B$, partially oriented 
$Ao->B$ and non-oriented $Ao-oB$, as well as some local constraint 
information $A*-\underline{*B*}-*C$%\\
 meaning that edges between A and B and 
between B and C cannot meet head to head at B. (Subsequently an asterisk (*) 
means any orientation of an edge end: e.g. $A*->B$ means either $A->B$ or 
$Ao->B$ or $A<->B$).%\\

 In a partially oriented including graph $\pi$:\\
\begin{itemize}
\item[(i)] A is a parent of B if and only if $A->B$ in $\pi$.%\\
\item[(ii)] B is a collider along the path $<A,B,C>$ if and only if 
$A*->B<-*C$ in $\pi$.%\\
\item[(iii)] An edge between B and A is into A iff $A<-*B$ in $\pi$%\\
\item[(iv)] An edge between B and A is out of A iff $A->B$ in $\pi$.%\\
\item[(v)] A is d-separated from B given set S iff A and B are conditionally 
independent given S.%\\
\item[(vi)] A and B are d-connected given C iff  there  exists  no 
such S 
containing C such that A and B are conditionally 
independent given S.%\\
\item[(vii)] In a partially oriented including path graph $\pi'$, U is a 
definite 
 discriminating path for B if and only if U is an undirected path between X 
and 
Y containing B, $B \neq X, B \neq Y$, every vertex on U except for B and the 
endpoints is a collider or a definite non-collider on U and:\\
(a) if V and V" are adjacent on U, and V" is between V and B on U, then 
$V*->V"$ on U,\\
(b) if V is between X and B on U and V is a collider on U, then $V->Y$ in 
$\pi$, else $V<-*Y$ on $\pi$\\
(c) if V is between Y and B on U and V is a collider on U, then $V->X$ in 
$\pi$, else $V<-*X$ on $\pi$\\
(d) X and Y are not adjacent in $\pi$.\\
\item[viii)] Directed path U: from X to Y: if V is adjacent to X on U then 
$X->V$ in $\pi$, if $V$ is adjacent to Y on V, then $V->Y$, if V and V" are 
adjacent on U 
and V is between X and V" on U, then $V->V"$ in $\pi$.%\\
\end{itemize}%
%\\

{\noindent \bf The Causal Inference (CI) Algorithm:}\\
Input: Empirical joint probability distribution\\
Output: partial including path graph $\pi$.\\
\begin{itemize}%\\
\item[A)] Form the complete undirected graph Q on the vertex set V.%\\
\item[B)] if A and B are d-separated given any subset S of V, remove the edge 
between 
A and B, and record S in Sepset(A,B) and Sepset(B,A). %\\
\item[C)] Let F be the graph resulting from step B). Orient each edge 
o-o.  For each 
triple of vertices A,B,C such that the pair A,B and the pair B,C are each 
adjacent in F, but the pair A,C are not adjacent in F, orient $(C)$ A*-*B*-*C 
as 
$A*->B<-*C$ if and only if B is not in Sepset(A,C), and orient 
 A*-*B*-*C 
as $A*-\underline{*B*}-*C$ if and only if B is in Sepset(A,C).%\\
\item[D)] Repeat%\\
\begin{itemize}
\item[if] there is a directed path from A to B, and an edge  A*-*B, orient 
$(D_p)$ A*-*B 
as $A*->B$,%\\
\item[else if] B is a collider along $<A,B,C>$ in $\pi$, B is adjacent to D, 
and A and C are not d-connected given D, then orient $(D_s)$ $B*-*D$ as 
$B<-*D$ ,%\\
\item[else if] U is a definite discriminating path between A and B for M in 
$\pi$ and 
P and R are adjacent to M on U, and P-M-R is a triangle, then\\
if M is in Sepset(A,B) then M is marked as non-collider on subpath 
$P*-\underline{*M*}-R$\\
else $P*-*AM*-*R$ is oriented $(D_d)$ as $P*->M<-*R$,%\\
\item[else if] $P*-\underline{>M*}-*R$ then orient $(D_c)$ as $P*->M->R$.%\\
\item[until] no more edges can be oriented.%\\
\end{itemize}
\item[End of CI]%\\
\end{itemize}

\section{From CI Output to Belief Network}
%----------------------------------------------------\\
Let us imagine that we have obtained a partial including path graph from CI, 
and we want to find a Belief Network representing the joint probability 
distribution out of it. Let us consider the following algorithm:\\

\noindent
{\bf CI-to-BN Algorithm}\\
Input: Result of the CI algorithm (a partial including path graph)\\
Output: A belief network%\\
\begin{itemize}
\item[A)] Accept unidirectional and bidirectional edges obtained from CI.%\\
\item[B)] Orient every edge $Ao->B$ as $A->B$.%\\
\item[C)]  Orient edges of type $Ao-oB$ either as $A<-B$ or $A->B$ so as not 
to 
violate  $P*-\underline{*M*}-*R$ constraints.
\item[End of CI-to-BN]%\\
\end{itemize}

We claim that:
\begin{th}
(i) By the CI-to-BN algorithm, a belief network can always be obtained.\\
(ii) The obtained belief network keeps all the dependencies and
independencies of the original underlying including path graph.
\end{th}

The rest of this section provides a sketchy proof of the above theorem.

First, we shall notice, that hiding variables - while 
keeping the connections of original variables - introduces into an including 
path graph two kinds of new 
edges: (1) those stemming from a direct connection: unidirectional, keeping 
the acyclicity of the graph, (ii) common cause connections, introducing 
bidirectionality, Let us notice, that the dag of the original network (the 
intrinsic one) introduced a partial ordering of the (observed and unobserved) 
nodes.  If a new (uni)directed edge is introduced by variable hiding, then 
only 
if there existed a directed path between the nodes. 
Therefore, the original partial order is not violated by them. On the other 
hand, a newly introduced bidirectional edge indicates that there exists a 
hidden node such that there exist a directed path from this node to each of 
the now connected nodes in the original graph. So it is justified to replace 
 the introduced bidirectional edge by a parentless common cause and in this 
way 
also the partial ordering of the original graph is not violated. Hence the 
obtained including path graph is in fact a dag. Also,  all 
the dependencies and independences represented by this graph would be really 
 those which are present in the original graph under the assumption of 
variable 
hiding. There remains only one cosmetic question to be settled: What happens 
if a new bidirectional edge is introduced between nodes which were already 
connected within the original dag. Should we keep the information that there 
was an edge between the nodes in the original graph between them or can 
we drop it without affecting the dependence/independence information ? It 
happens that we can drop it because the procedure introduces necessary 
 additional edges around this connection (vital for incoming directed edges) 
in 
order to keep all the paths considered by d-separation intact. Hence it is 
 legitimate to draw a dag (called below the full hiding dag) consisting only 
of 
observable nodes with uni- 
and bi-directional edges the latter ones indicating that the nodes are not 
connected but have a hidden common cause. %\\

Now let us consider the following question: Let us consider a  mixed graph 
(MG) having all 
the unoriented edges of the original including path graph (here called FHD - 
Full Hiding Dag) and the CI partially oriented  including path graph
(here called CIG - CI graph). Let all the unidirected edges of FHD be 
unidirected the same way in MG, let all the edges bidirected by CI he 
 bidirected in MG. Let all partially directed edges of CI be unidirected in 
MG. Last not least, let all bidirectional FHD edges not oriented at all by 
Ci be left unoriented in MG. Now if there were no cycles in MG, then also 
the claim (i) of the above theorem would be valid. The proof of acyclicity of 
MG is not difficult, but laborious. An overview can be made in terms of 
Figures, from Fig.\ref{abbzwei} to Fig.\ref{abbelf}.  In this series of 
Figures it is demonstrated that no three edges of MG can form a cycle. \\
Within the CI algorithm, only orientation steps denoted as $D_p$, $D_s$ and 
$C$ can give rise to a partial orientation of a FHD-bidirectional edge within 
CIG. (Step $D_d$ immediately creates a bidirectional edge in CIG out of it). 
The Figures summarize the proof as follows: e.g. in Fig.\ref{abbvier} (a) 
overviews the general situation in the FHD when one edge (BA) is bidirectional 
and AC and CB are unidirectional and the edge DA caused in step $(C)$ of CI 
 orientation of BA and DA towards A. (b) and (c) follow the case when the 
edges 
 DA and AC are not bridged: as a result the edge BA is made biorriented in 
CIG. 
On the other hand, (d) and (f) deal with the case when DA and AC are bridged: 
then also BA grows bidirectional in CIG. So. Fig.\ref{abbdrei}-\ref{abbvier} show 
that in case of one FHD-bidirectional and two FHD-unidirectional edges no 
cycle in MG is possible. Fig.\ref{abbfuenf}-\ref{abbneun} demonstrate the same for 
two FHD-bi- and one FHDunidirectional edges, demonstrating, that both 
 bidirectional edges will be made bidirectional in MG if there were any risk 
of 
cyclicity during the CI-algorithm. Fig.\ref{abbzehn}-\ref{abbelf} are concened with 
potential cycles consisting of three FHD-bidirectional edges   \\
Within a longer trail of edges it is immediately visible, that there must 
exist at least one pair of neighboring edges (one of them bidirectional in 
FHD) which are "bridged" that is 
their ends not neighboring on the path are neighbors in the graph. Hence 
we have here a triangle which - due to facts proven earlier - cannot form a 
cycle in MG and at least two edges and at least one FHD-bidirectional are 
oriented correctly (that is as in FHD), hence  cannot participate 
also in a larger cycle. (This is clear if the "third" edge has also been 
oriented by $C$, $D_s$ or $D_p$ step. But we can easily check, that if it has 
been oriented by the $D_d$ step, then the other edges will be oriented 
prohibiting a long-run cycle.) This completes the proof of claim (i).%\\

As claim (ii) is concerned, we shall first notice that a situation like that 
of Fig.\ref{abbzwoelf} cannot happen in an including path graph, that is it 
is never possible, that along a path $AB_1...B_nC$ with head to head meetings 
at $B_i$ one edge outgoing from each $B_i$ points at A and there is some j 
such that an edge $B_jC$ is outgoing from $B_j$. Now, when 
orienting edges according to CI-to-BN algorithm, we can make two types of 
errors:(a) introduce a path which is not active (in terminology of 
\cite{Geiger:90}) in BN, but is actually active in FHD, and  (b) introduce 
a path which is active (in terminology of \cite{Geiger:90}) in BN,  but is 
actually not active (blocked) in FHD.  In case (a), we may have the structure 
of such a path as $...,D,B_n,...,B_1,A,C_1,...,C_m,E,...$ in Fig. 
\ref{abbdreizehn}, with node A set in BN erroneously 
active (to the left) or passive (to the right).
Let us assume that this is the shortest active path between the nodes of 
interest that is no subset of nodes on the erroneously active path can form
also an active path. Then in Fig.\ref{abbdreizehn}.a) and .b) there exists
no unioriented edge $D->B_i$ nor $E->C_j$, nor bidirectional edge 
$B_i<->C_j$ nor $D<->B_j$ nor $E<->C_j$  nor 
$D<->A$ nor $E<->A$,for any i,j. And additionally in 
Fig.\ref{abbdreizehn}.a) there exists no edge $D->A$ nor $E->A$. In 
Fig.\ref{abbdreizehn}.b) there exists no edge $D<-A$ nor $E<-A$. 
 But it can then be 
demonstrated, that the CI orients correctly nodes from D to $B_1$ and from E 
 to $C_1$, and then a definite discriminating path for A emerges, and the 
edges 
at A are oriented correctly, hence it is denied that an error may occur at 
A.%%\\

As error (b) is concerned, we can proceed in an analogous way, also assuming 
that we have to do with the shortest erroneously passive path. \\

This would then complete the proof of the Theorem. 

\input ITABB.tex

\section{Summary and Outlook}

Within this paper an algorithm of recovery of belief network structure from 
data has been presented and its correctness demonstrated. It relies 
essentially on exploitation of the result  of the known CI 
algorithm of Spirtes, Glymour and Scheines \cite{Spirtes:93}. The edges of  
partial including path graph, not oriented by CI, are oriented to form a 
directed acyclic graph. The contribution of this paper is to show that such 
an orientation of edges always exists without necessity of adding auxiliary 
hidden variables, and that this dag captures all dependencies and 
independencies of the intrinsic underlying including path graph.%\\

The CI-to-BN algorithm will suffer from the very same shortcomings as the CI 
algorithm, that is it is  tractable only for a small number of edges. It will 
be interesting task to examine the possibility of such an adaptation of the 
Fast CI algorithm \cite{Spirtes:93}. It may not be trivial as the product of 
CI differs from that of FCI \cite{Spirtes:93}. Another path of research would 
be to elaborate of version  of  CI-to-BN  assuming   only  with  a 
restricted number 
 of variables participating in a d-separation, which would also bind the 
exponential explosion of numerical complexity.%\\

 %\\
\end{document}